# Structural Damage Identification Using Artificial Neural Network and Synthetic data

Divya Shyam Singh[a], GBL Chowdary[a], D Roy Mahapatra[a]

[a]Department of Aerospace, Indian Institute of Sciences, Banglore, India

Email: divya.ssingh.mec11@itbhu.ac.in

This paper presents real-time vibration based identification technique using measured frequency response functions(FRFs) under random vibration loading. Artificial Neural Networks (ANNs) are trained to map damage fingerprints to damage characteristic parameters. Principal component statistical analysis(PCA) technique was used to tackle the problem of high dimensionality and high noise of data, which is common for industrial structures. The present study considers Crack, Rivet hole expansion and redundant uniform mass as damages on the structure. Frequency response function data after being reduced in size using PCA is fed to individual neural networks to localize and predict the severity of damage on the structure. The system of ANNs trained with both numerical and experimental model data to make the system reliable and robust. The methodology is applied to a numerical model of stiffened panel structure, where damages are confined close to the stiffener. The results showed that, in all the cases considered, it is possible to localize and predict severity of the damage occurrence with very good accuracy and reliability.

## 1.Introduction

Structural health monitoring to detect damages at the earliest stage is observed as an effective technique to improve safety and reduce cost across aerospace, civil and mechanical industries. Many damage detection methods have attempted to identify damage by solving an inverse problem, which inevitably needs an analytical model. However, often the construction of these analytical model requires considerable effort in building a mathematical framework with acceptable level of accuracy and reliability which makes these approaches less attractive. In order to circumvent this complexity, numerous artificial neural network techniques have been applied to structural health monitoring and damage detection[1], [2]. A summary of literature pertaining to the various methods for damage detection and health monitoring of structures based on changes in their measured dynamic properties is presented in this section. The methods are categorized based on the type of measured data used, and/or the technique used to identify the damage from the measured data.

Kudva [3] used a back propagation neural network to identify damage in a plate stiffened with a 4 x 4 array of bays. Damage was modeled by cutting holes of various diameters in the plate at the centers of the bays. The authors found that neural network was able to predict the exact location of damage but the severity of damage was more problematic. Wu, et al. (1992) [4] used a back propagation neural network on a three-story building driven by earthquake excitation. Spillman, et al. [5] used feed- forward neural network to identify damage in a steel bridge element. In a similar study Rhim and Lee (1994)[6] used back propagation neural network to identify delamination damage in a composite cantilever beam. A lot of effort to train the neural network has been through the use of Frequency response functions. Good results have been obtained in the damage identification of numerically modeled structures[7]. A 'bottleneck' limiting the use of FRFs is the huge size of the required data set. Thus, one of the main challenges in FRF based damage identification is the development of algorithms that assist in the processing of the enormous amounts of FRF data. [7]–[11] present the use of FRF data for damage detection in different types of numerical and experimental structures; in some cases, techniques have been tested on real structures. The proposed damage identification method is based on FRF data. FRF data is one of the easiest to obtain in real time as they require only a small number of sensors and very little human involvement[12]. Measured FRF data are usually the most compact form of data obtained from vibration tests of structures. They provide an abundance of information on the structure dynamic behavior at master degrees of freedom over a range of frequency interest, whereas modal data are only subsets of FRF data [13]. Compared to modal parameter methods, FRF based damage detection techniques do not require extensive post data analysis.

In this proposed damage identification method, the direct FRF data is utilized as damage fingerprints to identify defect. Pattern changes in FRF data are analyzed by ANNs that map unique damage fingerprints to locations and severities of damage. To obtain suitable input data for network training and to filter uncertainties such as measurement noise the FRFs are compressed to few PCs using Principal Component Analysis, a statistical technique. The random Gaussian white noise signal is used as vibration input force signal to simulate real time conditions on the aircraft. The number of such signals is limited to 10 numbers to limit computational time. The FRFs are fed to their respective Neural Network to identify the location and severity of damage. The proposed method is verified on numerical and stiffened panel structure. An issue of limited sensor availability is considered on the structure.

## 2. Dimensionality Reduction

### 2.1 PCA of the Frequency Response Function

Principal Component Analysis(PCA) [14], [15] is a statistical procedure which allows identifying the principal directions in which the data varies. A key application of PCA is to reduce the dimensionality of the problem in the case where eigenvalues cover a wide dynamic range. In particular it allows identifying the principal directions in which the data varies. Principal components are entirely equivalent to finding the eigenvectors of the covariance matrix: each Eigenvector gives one principal component. The corresponding eigenvalue depicts the variance of that principal component, and the principal components with largest variances are the most important.

To produce damage indices that are feasible for neural network training, the size of the damage quantities (FRFs) must be greatly reduced. For the stiffened panel structure, a full size FRF, which covers a frequency range of 0 to 1000 Hz, contains 2,00,000 spectral lines. This would mean 2,00,000 input nodes in the neural network for each measurement point. Such large numbers of input points cause severe problems in training convergence. Therefore, PCA is applied to the damage fingerprints to reduce size and filter noise. The '*princomp*' function in MATLAB is utilized to project the damage fingerprints onto their PCs. The no of measurement points on the stiffened panel are 8, which include four tri-axial accelerometers and 4 Uni-directional strain gauges. Tri-axial accelerometer gives translational acceleration at the point of measurement in all three Cartesian directions and Uni-directional strain gauges measure strain in the direction of its length. Therefore, the total no of measurement points summing up to 16 (12 accelerations, 4 strains). For damage scenario, the measurement matrix would be of dimension 2,00,000 × 16. Such huge amount of data is reduced into 100 × 1 damage vector with use of Principal Component Analysis. The most significant Eigen values and its Eigenvectors are considered and remaining is discarded. The first seven principal values contribute to 99.999% of variability in data, thus projecting huge measurement data 2,00,000 × 16 onto 100 principal components, where first 84 components relate to the accelerance FRF and last 16 correspond to strain FRF.

### 2.2 Damage fingerprint in the reduced FRF

The main idea of data dimensionality reduction is its easiness to train Artificial Neural Network and have good convergence and also to filter out noise in the data. The dimensionality reduction using PCA will fingerprint the damage scenario to the reduced data. The figures 1 below show principal components variation for different damage scenarios.

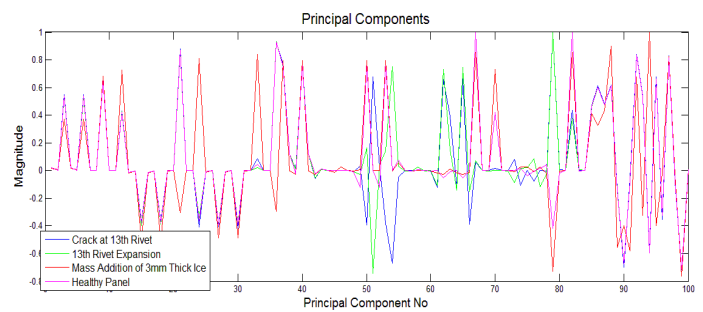

Figure 1: Principal components for crack damage at different locations

In general, the determination of the optimal number of PCs is dependent on the quality of damage patterns embedded in a data set as well as the level of noise pollution. As a general guideline, indications on dominant features of a data set are given by individual and cumulative contributions of PCs. For example, if the first 7 PCs account for more than 99% of the information of the original data, then a selection of more than 7 PCs for damage identification is probably unnecessary as the information gained by including higher PCs is negligible.

## 3. Neural Network

Artificial neural networks (ANNs) are derivative of biological neural system. This research was driven by the desire to build better pattern recognition and information processing system[16]. The most popular class of multi- layer feed-forward neural networks is multi-layer perceptron in which each computational unit employs either the threshold function or the sigmoid function. Multilayer perceptron can form arbitrarily complex decision boundaries and represent any Boolean function. The development of the back-propagation learning algorithm to determine weights has made these networks the most popular among the researchers and users of neural networks. The Multi-Layer-Perceptron was first introduced by [17]. The figure 2 shows a typical 3 layer perceptron network

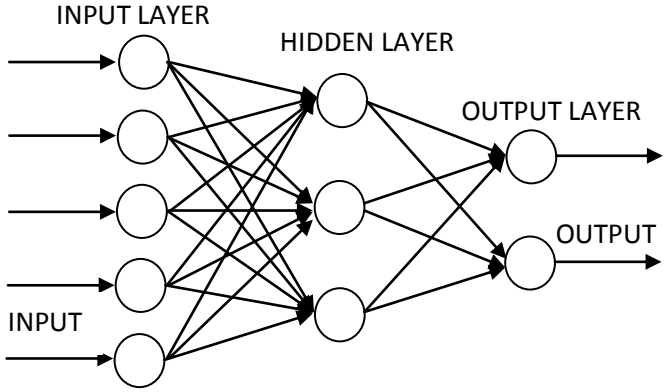

Figure 2. Typical 3 layer neural network

Let $\{(h^1,d^1),(h^2,d^2)\ldots\ldots,(h^p d^p)\}$ be a set of p training patterns(input-output) where $h^i \in R^n$ (the elements of FRF matrix) is the input vector in the n-dimensional pattern space, and $d^i$, an output vector. The squared error cost function most frequently used in the ANN literature is defined as:

$$E = \frac{1}{2}\sum_{i=1}^{p}\|y^i - d^i\|^2 \qquad (7)$$

where $y^i$ is the desired outputs. The backpropagation algorithm is a gradient-decent method to minimize the above squared error cost function. The backpropagation algorithm falls into the class of supervised learning, where a learning algorithm is trained on examples of desired behavior. In a backpropagation multi-layer network, the outputs of one layer become the input to the subsequent layer.

*Step 1*: Initialize the weights are initialized a small random values.

*Step 2*: A training set is randomly chosen and fed into the neural network. For each training pair, steps 3-8 are repeated.

*Step 3*: The first layer receives input signal $h^i$ and sends this to the adjacent layer(the hidden unit). Each hidden unit($z_j, j = 1,\ldots\ldots m$) sums its weighted input signals as

$$(z_{in})_j = (w_0)_j + \sum_{i=1}^{p} x_i (w_i)_j \qquad (8)$$

This acts as an input for the activation function to compute its output signal, $z_j = \theta\big((z_{in})_j\big)$ and sends this signal to all units to the next hidden layer.

*Step 4*: Each output unit $(y_k, k = 1,\ldots\ldots,n)$ sums its weighted input signals as

$$(y_{in})_k = (w_0)_k + \sum_{j=1}^{m} z_j (w_j)_k \qquad (9)$$

And applies its activation function to compute its output signal, $y_k = \theta((y_{in})_k)$.

*Step 5*: Each output unit $(y_k, k = 1,\ldots\ldots,n)$ receives a target pattern corresponding to the input training pattern, compute error information term

$$\delta_k = (d_k - y_k)\theta^{'}(y_{in})_k \qquad (10)$$

Calculate its weighted and bias correction term(used to update $w_{jk}$ later)

$$\Delta w_{jk} = \alpha \delta_k z_j \qquad (11)$$

$$\Delta w_{0k} = \alpha \delta_k \qquad (12)$$

*Step 6*: Each hidden unit $(z_j, j = 1,\ldots\ldots m)$ sums its delta inputs and multiplies by the derivative of its activation function to calculate its error information term,

$$(\delta_{in})_j = \sum_{i=1}^{n} \delta_k w_{jk} \qquad (13)$$

$$\delta_j = (\delta_{in})_j \theta^{'}((z_{in})_j) \qquad (14)$$

Calculate its weight and bias term

$$\Delta w_{ij} = \alpha \delta_j x_j \qquad (15)$$

$$\Delta w_{0j} = \alpha \delta_j \quad (16)$$

*Step 7*: Each output and hidden units $(y_k, k = 1, \ldots, n)$ update its bias and weights

$$\left(w_{jk}\right)_u = \left(w_{jk}\right)_o + \Delta w_{jk} \quad (17)$$

$$\left(w_{ij}\right)_u = \left(w_{ij}\right)_o + \Delta w_{ij} \quad (18)$$

## 4. Damage Identification through Neural Networks

The neural networks presented here is used to predict the location and severity of the damage. The compressed FRF data is fed to its respective neural network to identify the location and calculate the severity of damage. The stiffened panel consists of 34 rivets. The binary vector of size 34 × 1 characterizes the location of rivet where the damage occurred. The vector with all '0' denotes healthy structure. Bayesian Regularisation learning algorithm is used for the above neural network. The performance graphs for the Crack neural network is presented in Figures 3. The performance parameter monitored is the mean squared error, which is 0.021107 at the best network.

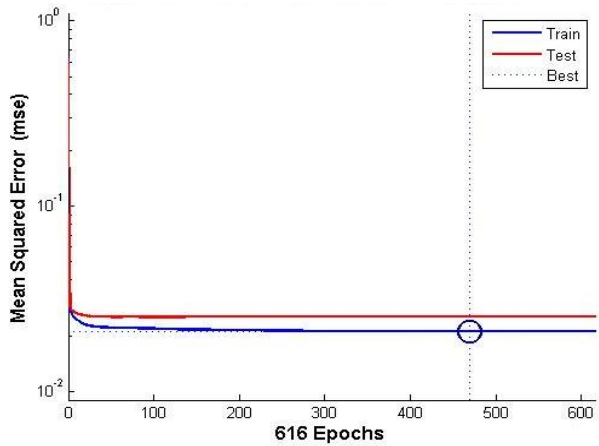

Figure 3 Performance graph of Crack damage localization

The numerical structure(Figure 4) containing crack is subjected to random loading time signal, the time response signal from 8 measurement locations are acquired, transformed to frequency domain using Fast Fourier Transform procedure to obtained frequency response functions. The huge FRF data is compressed to vector of 100 × 1 with Principal component analysis technique. To localize the crack damage, this FRF data is given as input to Crack localization neural network. The bar plot of the same is shown in the Figure 5. The neural network is unable to pinpoint to the exact location, but able to point Rivet 7 and Rivet 8 as locations of damage. To find the severity of the crack at the damage location, the data is again fed to severity neural network. The output shown in Figure 6 indicates a crack length of 10.88mm at the crack location 1 and a magnitude of 2.36 at location 2.

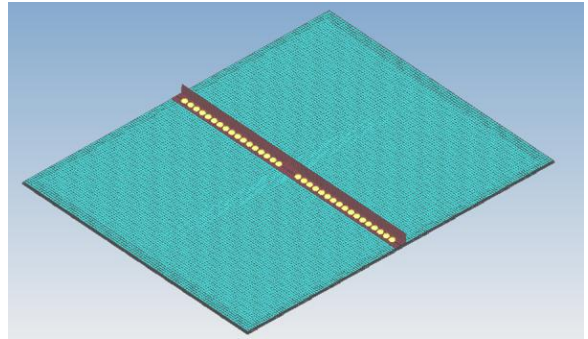

Figure 4 Stiffened Panel

The neural network is capable of learning the crack localization and parameter estimation. The percentage of misclassification on both neural networks is less than 20%. The neural network may not pinpoint the exact location at all times, but it is capable of directing to group of rivets as damaged. Later, the compressed data, after damage is localized, is keyed to severity prediction neural networks.

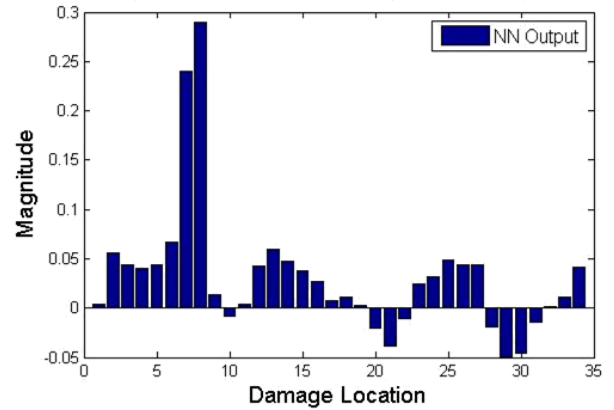

Figure 5 Crack damage locations on the panel

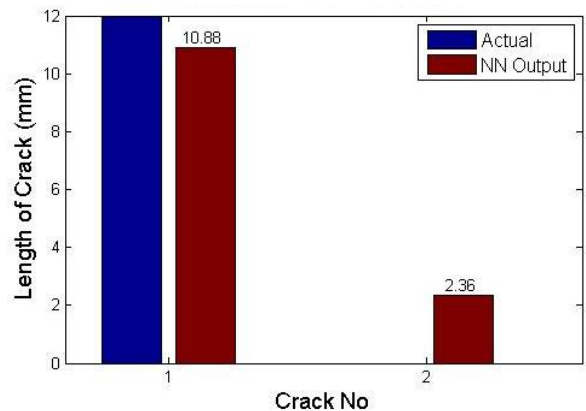

Figure 6 Crack Severity at the predicted location

## 5. Conclusion

This paper dealt with artificial neural networks (ANNs) in vibration based damage identification and their application in aerospace industry. The damage identification scheme proposed is based on frequency response functions (FRFs). The method uses neural network techniques and principal component analysis for damage feature extraction and noise reduction. To verify the proposed damage identification procedures, numerical studies were undertaken on stiffened panel structure. Field testing conditions were considered with limited no of sensors.

To overcome the obstacle of the large size of FRF data, the implementation of PCA techniques was suggested. Besides data reduction, PCA also offers the benefit of noise reduction and damage feature extraction, which further assists to reduce uncertainties from sources such as measurement noise and environmental fluctuations. The principal component analysis proved to be an efficient technique to compress the data without any loss of information of damage fingerprints. To investigate the performance of the proposed FRF based damage identification method, it was applied to the stiffened panel structure. The proposed FRF based damage identification method proved to be accurate and robust in the damage identification on stiffened panel structure. Crack localization neural networks can either pinpoint location or guide to group of rivets indicating damage. The percentage of misclassification is less than 15%. Exact prediction of severity of damage is a complex problem and more training data is required to predict accurately. The severity prediction neural network is still good to approximate severity of damage within an error percentage of 30-35%. Different learning rules are used for different neural networks to obtain better performance. The Bayesian regularization learning rule produce good neural network even for a small training set data, but at an expense of computational power. It was observed that re-training artificial neural networks several times will improve its performance characteristics.